\documentclass[lettersize,journal]{IEEEtran}
\usepackage{array}
\usepackage{textcomp}
\usepackage{stfloats}
\usepackage{url}
\usepackage{verbatim}
\usepackage{graphicx}
\usepackage{cite}
\usepackage{booktabs}  
\usepackage{subcaption} 
\usepackage{multirow}   
\usepackage{algorithm}
\usepackage{algpseudocode}  
\usepackage{amsmath}
\usepackage{amssymb}
\begin{document}

\title{SAM-Aug: Leveraging SAM Priors for Few-Shot Parcel Segmentation in Satellite Time Series}

\author{Kai~Hu, 
        Yaozu~Feng, 
        Vladimir~Lysenko, 
        Ya~Guo~\IEEEmembership{Member,~IEEE}, 
        and~Huayi~Wu~\IEEEmembership{Member,~IEEE}
        \thanks{Manuscript received ; revised .}
        \thanks{K. Hu, Y. Feng, and Y. Guo are with the Key Laboratory of Advanced Process Control for Light Industry, Ministry of Education, Jiangnan University, Wuxi 214122, China (e-mail: hukai\_wlw@jiangnan.edu.cn, fengyaozu19991105@163.com, guoy@jiangnan.edu.cn).}
        \thanks{V. Lysenko is with Academy of Biology and Biotechnology, Southern Federal University, Rostov-on-Don 344041, Russia (e-mail: vs958@yandex.ru).}
        \thanks{H. Wu is with the State Key Lab of Information Engineering in Surveying, Mapping and Remote Sensing, Wuhan University, Wuhan 430079, China (e-mail: wuhuayi@whu.edu.cn).}
       }

\maketitle

\begin{abstract}
Few-shot semantic segmentation of time-series remote sensing images remains a critical challenge, particularly in regions where labeled data is scarce or costly to obtain. While state-of-the-art models perform well under full supervision, their performance degrades significantly under limited labeling, limiting their real-world applicability. In this work, we propose \textbf{SAM-Aug}, a new annotation-efficient framework that leverages the geometry-aware segmentation capability of the Segment Anything Model (SAM) to improve few-shot land cover mapping. Our approach constructs cloud-free composite images from temporal sequences and applies SAM in a fully unsupervised manner to generate geometry-aware mask priors. These priors are then integrated into training through a proposed \textbf{RegionSmoothLoss}, which enforces prediction consistency within each SAM-derived region across temporal frames, effectively regularizing the model to respect semantically coherent structures. Extensive experiments on the PASTIS-R benchmark under a 5\% labeled setting demonstrate the effectiveness and robustness of SAM-Aug. Averaged over three random seeds (42, 2025, 4090), our method achieves a mean test mIoU of \textbf{36.21\%}, outperforming the state-of-the-art baseline by \textbf{+2.33\%} — a relative improvement of \textbf{6.89\%}. Notably, on the most favorable split (seed=42), SAM-Aug reaches a test mIoU of \textbf{40.28\%}, representing an 11.2\% relative gain with no additional labeled data. The consistent improvement across all seeds confirms the generalization power of leveraging foundation model priors under annotation scarcity.Our results highlight that vision models like SAM can serve as useful regularizers in few-shot remote sensing learning, offering a scalable, plug-and-play solution for land cover monitoring without requiring manual annotations or model fine-tuning.
\end{abstract}


\section{Introduction}
\IEEEPARstart{S}{emantic} segmentation of time-series remote sensing images plays a pivotal role in land cover monitoring, urban planning, and environmental change detection. Recent advances in deep learning, particularly vision transformers and multi-scale fusion architectures, have relatively improved the accuracy of pixel-level classification by modeling both spatial and temporal dynamics. Models such as Exchanger combined with mask2former have established state-of-the-art performance on benchmarks like PASTIS, achieving high mIoU under full supervision. However, these models typically require large amounts of densely annotated training data—a condition that is often impractical in real-world scenarios due to the high cost and labor intensity of expert labeling.

This limitation becomes particularly acute in few-shot learning settings, where only a small fraction of labeled samples are available. Under such constraints, even advanced models suffer from severe performance degradation, as evidenced by our experiments: when trained on only 5\% of the PASTIS-R dataset, the current SOTA model achieves a test mIoU of 33.88\%, a substantial drop from 69.7\% under full supervision. This performance gap highlights a critical challenge: how to enhance model generalization and robustness when labeled data is extremely scarce.

To address this issue, we explore the use of \textbf{geometry-aware mask  priors} derived from foundation models to regularize training in low-data regimes. Specifically, we leverage the recently introduced \textbf{Segment Anything Model (SAM)}~\cite{kirillov2023segment}, which demonstrates strong geometry-aware segmentation capabilities across diverse domains. Unlike traditional data augmentation or self-supervised methods that rely on geometric transformations or pretext tasks, SAM provides rich, semantically meaningful region structures without requiring any task-specific annotations. In this work, we propose \textbf{SAM-Aug}, a novel augmentation framework that integrates SAM-generated masks into the training pipeline of time-series semantic segmentation models.

Our key insight is that regions identified by SAM—despite being derived from a single cloud-free composite—capture coherent land cover patterns that should exhibit consistent semantic predictions across temporal frames. To exploit this prior, we introduce the \textbf{RegionSmoothLoss}, a consistency regularization term that minimizes the variance of predicted class probabilities within each SAM-derived region over time. This encourages the model to produce stable and semantically smooth predictions, effectively reducing noise and overfitting to limited labels.

We evaluate SAM-Aug on the PASTIS-R dataset under a challenging 5\% labeled setting, using the same training, validation, and testing splits for fair comparison with the baseline. Experimental results show that our method achieves a \textbf{test mIoU of 36.21\%}, outperforming the current SOTA by \textbf{+2.33\%}—a relative improvement of 6.89\%—while maintaining comparable inference speed and training efficiency. Furthermore, ablation studies and visualizations confirm the effectiveness of both the SAM-based region prior and the proposed loss function.

In summary, this work makes the following contributions:  
(1) We propose \textbf{SAM-Aug}, a novel few-shot learning framework that leverages geometry-aware segmentation priors from SAM for time-series remote sensing image analysis.  
(2) We introduce \textbf{RegionSmoothLoss}, a simple yet effective consistency loss that enforces intra-region prediction stability across temporal frames.  
(3) We demonstrate relative performance gains under extreme label scarcity, providing a practical and annotation-efficient solution for real-world land cover mapping.

\section{Related Work}
\subsection{Few-Shot Semantic Segmentation in Remote Sensing}
Few-shot learning has gained increasing attention in remote sensing due to the high cost of pixel-level annotations. Early approaches adapted meta-learning frameworks such as Prototypical Networks~\cite{snell2017prototypical} to the remote sensing domain, training models to rapidly adapt to novel classes with only a few labeled examples~\cite{jia2025generalized}. More recent methods leverage pre-trained vision transformers~\cite{touvron2021training} or contrastive learning~\cite{chen2020simple} to extract transferable features from large unlabeled datasets~\cite{lee2024unlocking, li2021scl}. However, these methods often rely on strong assumptions about class similarity or require auxiliary datasets for pre-training. In contrast, our work operates in a realistic setting where only a small fraction of labels from the target domain are available, and we introduce an annotation-free prior from a foundation model to enhance generalization without task-specific pre-training.

\subsection{Foundation Models in Remote Sensing}
The emergence of foundation models—trained on massive, diverse datasets—has opened new avenues for data-efficient learning. Models like CLIP~\cite{radford2021learning} and DINOv2~\cite{oquab2023dinov2} have been adapted for zero-shot classification and semantic segmentation in remote sensing by aligning image features with text prompts or clustering in self-supervised feature spaces~\cite{bou2024exploring, stacchio2025rsplitzero}. Recently, the Segment Anything Model (SAM)~\cite{kirillov2023segment} has demonstrated better geometry-aware capability in generating high-quality segmentation masks across natural images. Several works have explored SAM's potential in medical imaging~\cite{ali2025review}, agriculture~\cite{luo2023survey}, and urban scene understanding~\cite{rehman2024effective}, often fine-tuning SAM on downstream tasks or using it as an annotation assistant. In the remote sensing community, preliminary studies have evaluated SAM's performance on aerial imagery~\cite{osco2023segment}, showing moderate success but highlighting domain gaps due to the unique spectral, spatial, and temporal characteristics of satellite data. Our work differs in that we do not fine-tune SAM nor use its masks as direct supervision; instead, we treat SAM-generated regions as soft priors to regularize model training through temporal consistency, enabling plug-and-play integration without any adaptation.

\subsection{Consistency Regularization and Temporal Smoothing}
Consistency-based learning has proven effective in semi-supervised and low-data regimes by enforcing invariant predictions under perturbations~\cite{wang2024towards}. In time-series remote sensing, temporal consistency is a useful prior: land cover types exhibit stable spatial structures over time despite seasonal variations and atmospheric conditions. Methods such as Temporal Ensembling~\cite{laine2016temporal} and Mean Teacher~\cite{tarvainen2017mean} have been adapted to encourage stable predictions across temporal frames~\cite{araslanov2021self}. More recent works introduce smoothness constraints at the feature or output level to reduce noise and improve segmentation coherence~\cite{lan2023smooseg}. Our \textbf{RegionSmoothLoss} builds upon this idea but introduces a key innovation: rather than applying smoothing uniformly or based on heuristic neighborhoods, we use SAM-derived regions as semantic anchors to define meaningful spatial support. This allows the model to preserve fine-grained boundaries while promoting intra-region consistency, helping to leverage the geometry-aware mask prior from a foundation model to guide learning under label scarcity.

\section{Methodology}
\subsection{Overview}
\begin{figure*}[htbp]
    \centering
    \includegraphics[width=0.9\linewidth]{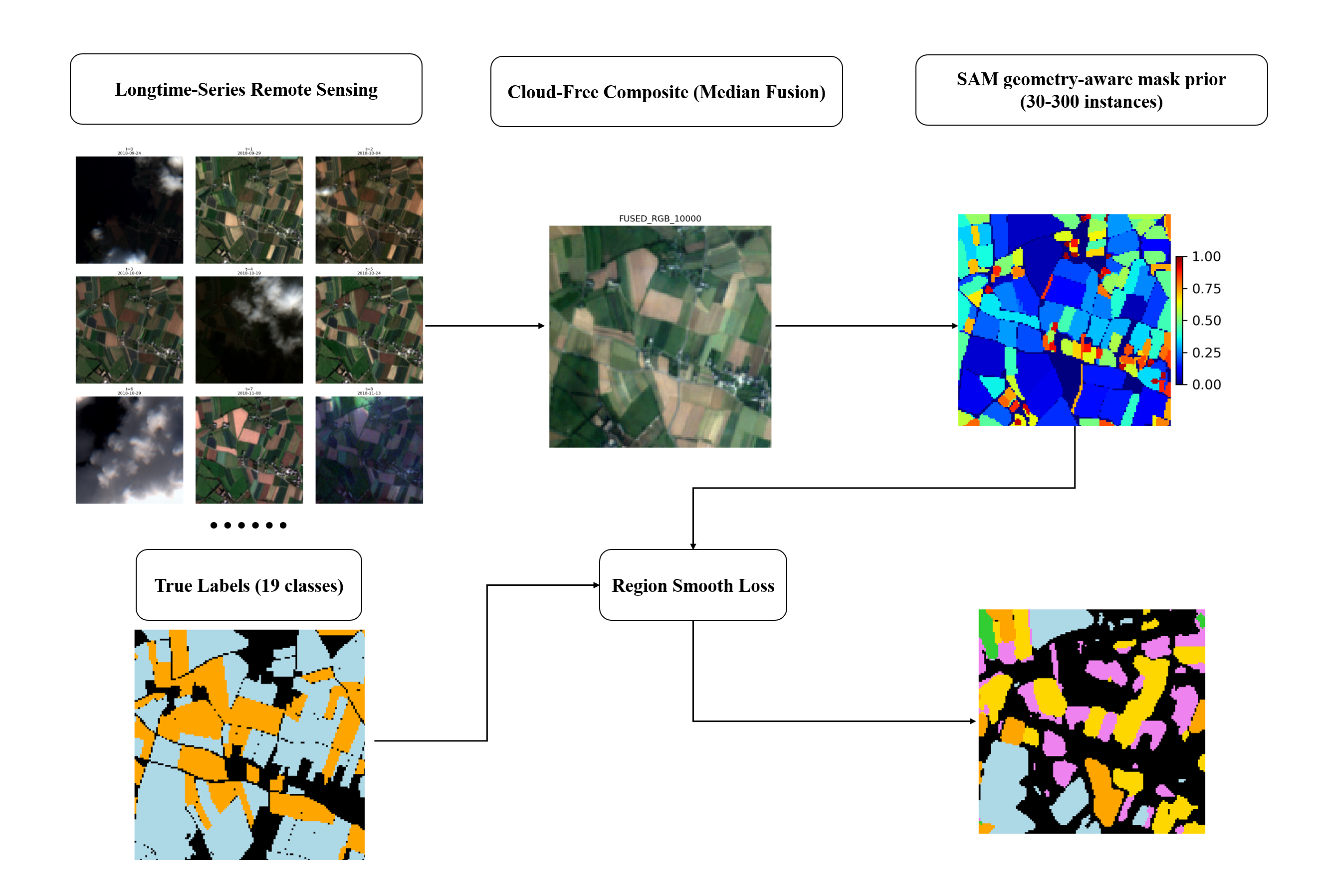}
    \caption{Overview of the proposed SAM-Aug framework. A cloud-free composite is generated from multi-temporal inputs, SAM extracts region masks, and RegionSmoothLoss enforces prediction consistency within each region across time.}
    \label{fig:pipeline}
\end{figure*}

The proposed framework, \textbf{SAM-Aug}, aims to enhance few-shot semantic segmentation of time-series remote sensing images by leveraging geometry-aware  mask priors from the Segment Anything Model (SAM). As illustrated in Figure~\ref{fig:pipeline}, our method consists of three key stages: (1) generation of cloud-free composite images from multi-temporal sequences, (2) automatic extraction of semantic region masks using SAM, and (3) integration of these masks into the training process via a novel \textbf{RegionSmoothLoss} that enforces temporal consistency within each region. The backbone model (e.g., Exchanger + mask2former) remains unchanged, ensuring compatibility with existing architectures. The entire framework operates under a limited-label setting, where only a small percentage of training samples are annotated.

\subsection{Cloud-Free Composite and SAM-Based Region Proposal}
To obtain a clean and representative reference image for region extraction, we first generate a cloud-free composite from the input time series. Given a sequence of $T$ multi-spectral images $\mathcal{X} = \{x_t \in \mathbb{R}^{H \times W \times C}\}_{t=1}^T$, we select frames with minimal cloud coverage and compute their pixel-wise median or mean:
\begin{equation}
x_{\text{composite}} = \frac{1}{|S|} \sum_{t \in S} x_t,
\end{equation}
where $S \subseteq \{1,\dots,T\}$ is the set of cloud-free or low-cloud frames.

\begin{equation}
\label{eq:cloud_detection}
\mathcal{C}_t = 
\left( B2_t > 0.2 \right) \cap 
\left( B8_t > 0.3 \right) \cap 
\left( B12_t > 0.2 \right) \cap 
\left( \mathrm{NDWI}_t < 0.9 \right)
\end{equation}

We then feed $x_{\text{composite}}$ into the pre-trained Segment Anything Model (SAM) to generate semantic-agnostic but spatially coherent region masks $\mathcal{M}_{\text{SAM}} = \{m_i \in \{0,1\}^{H \times W}\}_{i=1}^Q$, where $Q$ is the number of regions (typically 30–300). SAM is used without fine-tuning.

\begin{algorithm}[t]
\caption{SAM Prior Generation from Satellite Image Time Series}
\label{alg:sam_prior_generation}
\begin{algorithmic}[1]
\Require
    Sample ID $i$, S2 data path $\mathcal{P}$,
    SAM checkpoint $\theta_{\text{SAM}}$, output directory $\mathcal{D}$
\Ensure
    SAM-derived spatial prior $\mathbf{S}_i \in \mathbb{Z}^{H \times W}$, fused RGB $\mathbf{R}_i \in [0,1]^{3 \times H \times W}$

\State $\mathbf{X} \gets \text{LoadS2Data}(\mathcal{P}, i)$ \Comment{$[T, C, H, W]$}
\State $\mathcal{T}_{\text{cloud}} \gets \text{DetectClouds}(\mathbf{X})$ \Comment{Eq.~\eqref{eq:cloud_detection}}
\State $\mathbf{b} \gets \text{MeanIntensity}(\mathbf{X})$, $\mathbf{s} \gets \text{SharpnessScore}(\mathbf{X})$
\State $\mathcal{T}_{\text{clear}} \gets \{ t \mid \mathcal{T}_{\text{cloud}}[t] < 0.8 \}$
\If{$|\mathcal{T}_{\text{clear}}| = 0$}
    \State $\mathcal{T}_{\text{clear}} \gets \{\arg\min_t \mathcal{T}_{\text{cloud}}[t]\}$
\EndIf
\State $t^* \gets \arg\max_{t \in \mathcal{T}_{\text{clear}}} \left( -10 \cdot \mathcal{T}_{\text{cloud}}[t] + \mathbf{b}[t] + \mathbf{s}[t] \right)$
\State $\mathbf{R}_{\text{fused}} \gets \text{WeightedFuseRGB}(\mathbf{X}, \mathcal{T}_{\text{clear}})$
\State $\mathbf{R}_{\text{uint8}} \gets \text{Clip}(\mathbf{R}_{\text{fused}} \times 255, 0, 255)$
\State $\mathcal{M} \gets \text{SAM-AMG}(\mathbf{R}_{\text{uint8}}; \theta_{\text{SAM}})$ \Comment{Loose settings}
\State $\mathbf{S}_i \gets \text{CreateSegMap}(\mathcal{M})$ \Comment{Sorted by area}
\State $\text{Save}(\mathbf{S}_i, \mathcal{D}/\text{SAM\_PRIOR}\_i.\text{npy})$
\State $\text{Save}(\mathbf{R}_{\text{fused}}, \mathcal{D}/\text{FUSED\_RGB}\_i.\text{npy})$
\State \Return $\mathbf{S}_i, \mathbf{R}_{\text{fused}}, t^*$
\end{algorithmic}
\end{algorithm}

\subsection{RegionSmoothLoss for Temporal Consistency Regularization}
Let $\mathcal{Y} = \{y_t \in \mathbb{R}^{H \times W \times K}\}_{t=1}^T$ denote the model's raw predictions for $K$ land cover classes, and let $M \in \mathbb{Z}^{H \times W}$ be the SAM-derived region mask. The RegionSmoothLoss is defined as:

\begin{equation}
\mathcal{L}_{\text{region}} = \frac{1}{Q} \sum_{q=1}^{Q} \frac{1}{T} \sum_{t=1}^{T} \frac{1}{K} \sum_{k=1}^{K} \mathrm{Var}\left( \sigma(y_t)_{ijk} \mid (i,j) \in \mathcal{R}_q \right),
\end{equation}

where:
\begin{itemize}
    \item $q$ indexes the $Q$ regions,
    \item $\sigma(\cdot)$ denotes the sigmoid function,
    \item $\mathcal{R}_q$ denotes the set of pixel coordinates belonging to the $q$-th SAM-derived region, and $\mathrm{Var}(\cdot)$ computes the variance of predicted probabilities across pixels within $\mathcal{R}_q$ for each class $k$.
    \item $\mathrm{Var}(\cdot)$ computes the pixel-wise variance within the region.
\end{itemize}

The total training objective combines the segmentation loss and the regularization:
\begin{equation}
\mathcal{L}_{\text{total}} = \mathcal{L}_{\text{seg}} + \lambda \cdot \mathcal{L}_{\text{region}},
\end{equation}
where $\lambda$ controls the strength of spatial-temporal smoothing. We set $\lambda = 50$ based on a sensitivity analysis (see Section~\ref{sec:ablation}), which shows that this value yields optimal performance while maintaining stable regularization.

\begin{algorithm}[t]
\caption{RegionSmoothLoss: Region-Aware Smoothness Regularization}
\label{alg:region_smooth_loss}
\begin{algorithmic}[1]
\Require
    Predicted class probabilities $\mathbf{P} \in \mathbb{R}^{B \times K \times H \times W}$,
    SAM-generated region masks $\mathbf{S} \in \mathbb{Z}^{B \times H \times W}$
\Ensure
    Scalar loss $\mathcal{L}_{\text{region}}$

\State $\mathcal{L}_{\text{region}} \gets 0$, $N \gets 0$ \Comment{Initialize loss and region counter}
\State $\mathbf{P}_{\text{prob}} \gets \sigma(\mathbf{P})$ \Comment{Apply sigmoid: $\sigma(\cdot)$ maps logits to $[0,1]$}
\If{$\mathbf{S}.\text{shape}[-2:] \neq (H, W)$}
    \State $\mathbf{S} \gets \text{Interpolate}(\mathbf{S}, \text{size}=(H, W), \text{mode='nearest'})$
\EndIf

\For{each $b \in \{1, \dots, B\}$}
    \State $\mathbf{s}_b \gets \mathbf{S}[b].\text{flatten}()$ \Comment{Flatten SAM mask: $[H \times W]$}
    \State $\mathbf{p}_b \gets \mathbf{P}_{\text{prob}}[b].\text{reshape}(K, -1)$ \Comment{Reshape preds: $[K, H \times W]$}
    \State $\mathcal{U}_b \gets \text{Unique}(\mathbf{s}_b)$ \Comment{Get unique region IDs in batch $b$}

    \For{each $u \in \mathcal{U}_b$}
        \State $\mathbf{m}_u \gets (\mathbf{s}_b == u)$ \Comment{Binary mask for region $u$}
        \State $n_u \gets \|\mathbf{m}_u\|_1$ \Comment{Number of pixels in region $u$}
        \If{$n_u < 2$}
            \State continue
        \EndIf \Comment{Skip regions with fewer than 2 pixels}

        \State $\mathbf{v}_u \gets \mathbf{p}_b[:, \mathbf{m}_u]$ \Comment{Extract probs for region $u$: $[K, n_u]$}
        \State $\boldsymbol{\sigma}_u^2 \gets \mathrm{Var}(\mathbf{v}_u, \text{dim}=1)$ \Comment{Variance per class: $[K]$}
        \State $\mathcal{L}_{\text{region}} \gets \mathcal{L}_{\text{region}} + \frac{1}{K} \sum_{k=1}^{K} \sigma_{u,k}^2$
        \State $N \gets N + 1$
    \EndFor
\EndFor

\If{$N > 0$}
    \State $\mathcal{L}_{\text{region}} \gets \mathcal{L}_{\text{region}} / N$
\Else
    \State $\mathcal{L}_{\text{region}} \gets \text{zeros}(1).\text{requires\_grad\_()}$
\EndIf

\State \Return $\mathcal{L}_{\text{region}}$
\end{algorithmic}
\end{algorithm}

\subsection{Training Protocol}
We follow a 5\% labeled setting (seed=42), with 10\% for validation and 100\% for testing. SAM masks are pre-computed and fixed. The model is trained using AdamW with cosine decay and gradient clipping.

\subsection{Data Preprocessing and Augmentation Strategy}
Unlike natural image segmentation tasks where geometric transformations such as rotation, flipping, and scaling are commonly used for data augmentation, we avoid applying such operations in our framework due to the spatio-temporal coherence requirements of satellite time series. 
Preserving the original spatial structure and temporal order is critical for modeling phenological patterns and land cover dynamics.

Instead, we adopt two lightweight strategies to improve model robustness without introducing unrealistic distortions:

(1) \textbf{Random spatial cropping}, which increases diversity by sampling different sub-regions within each field;
(2) \textbf{Random temporal dropout}, where a random subset of time steps is dropped during training to simulate missing observations (e.g., due to cloud cover) and encourage temporal consistency.

All inputs are normalized using channel-wise mean and standard deviation computed from the training folds, and pixel values are clipped to $[0, 65535]$ to handle sensor saturation.

\section{Experiments}
\label{sec:experiments}

In this section, we present a comprehensive evaluation of the proposed SAM-Aug framework on the PASTIS-R dataset under a challenging few-shot setting. We first describe the experimental setup, including dataset configuration, evaluation metrics, and implementation details. Then, we report main results comparing our method with strong baselines, followed by ablation studies to validate the effectiveness of each component. Finally, qualitative visualizations are provided to illustrate the advantages of leveraging geometry-aware mask priors.

\subsection{Dataset and Evaluation Metrics}
We evaluate all methods on the PASTIS-R dataset~\cite{garnot2021panoptic,garnot2021mmfusion}, a large-scale benchmark for semantic segmentation of multi-temporal satellite imagery. PASTIS-R contains 42,390 agricultural parcels across France, with pixel-wise annotations for 19 land cover classes (including 18 crop types and one "other" class). The data includes both Sentinel-1 (S1) and Sentinel-2 (S2) observations over a 2-year period, resulting in sequences of up to 48 time steps per patch.

We follow the official 5-fold cross-validation split. For the few-shot setting, we use only 5\% of labeled training samples (randomly sampled with seed 42), while using 10\% for validation and full labels for testing. This simulates real-world scenarios where annotation effort is severely limited.

The primary evaluation metric is mean Intersection-over-Union (mIoU), which measures per-class segmentation accuracy and is widely adopted in remote sensing. We also report Overall Accuracy (OA) for completeness. All metrics are computed on the test set.

\subsection{Implementation Details}
Our implementation is based on the official codebase of \cite{cai2023rethinking}. The backbone model is \textbf{Exchanger}~ with a Mask2Former-style decoder\cite{cheng2022masked}. We use AdamW optimizer with initial learning rate $1\times10^{-4}$, weight decay $1\times10^{-2}$, and cosine learning rate decay. The batch size is set to 2 different from the official setting where the batch size is set to 1, and models are trained for 100 epochs and validation mIoUs are computed for every epoch. Input patches are cropped to $64 \times 64$ pixels via random spatial cropping, and temporal dropout is applied with rate $[0.1, 0.3]$ during training.

For SAM-Aug, we use the \texttt{vit-huge} variant of SAM~\cite{li2023segment} with point grid prompting to generate region masks. The cloud-free composite is formed by taking the median of all available S2 bands across time. SAM is used  without fine-tuning. 

All experiments are conducted on a single NVIDIA P100 GPU (16GB). SAM mask generation is performed once offline, adding negligible overhead to training.

\subsection{Baseline Architecture and Full-Data Context}
\label{sec:baseline_architecture}

To ensure a rigorous evaluation of data efficiency, we first assess the performance of several state-of-the-art models under full supervision (100\% training data) on the PASTIS dataset, as summarized in Table~\ref{tab:full_data_comparison}. Among these, the \textit{Exchanger+mask2former} architecture~\cite{cai2023rethinking} achieves the highest overall accuracy (OA: 83.46\%) and mean Intersection-over-Union (mIoU: 67.90\%), demonstrating superior modeling capacity for spatio-temporal remote sensing sequences.

\begin{table*}[t]
\centering
\caption{Performance comparison of existing methods on the PASTIS dataset under full-data setting (100\% training data). The best results are in bold.}
\label{tab:full_data_comparison}
\begin{tabular}{l||cccccc|cc}
\hline
Method & Beet & W.rapeseed & Corn & Orchard & Mixed cereal & Sorghum & OA(\%) & mIoU(\%) \\
\hline
Unet-3D~\cite{rustowicz2019semantic} & 93.5 & 92.1 & 94.1 & 50.5 & 44.7 & 36.5 & 81.32 & 58.30 \\
FPN~\cite{martinez2021fully} & 92.4 & 94.8 & 94.1 & 38.5 & 46.7 & 23.5 & 81.62 & 58.29 \\
UconvLSTM~\cite{russwurm2018convolutional} & 96.2 & 92.9 & 92.9 & 61.2 & 47.6 & 37.1 & 82.38 & 59.91 \\
BuconvLSTM~\cite{martinez2021fully} & 95.3 & 95.3 & 94.5 & 67.7 & 52.8 & 33.6 & 81.88 & 58.50 \\
TSViT~\cite{tarasiou2023vits} & 96.3 & 93.8 & 95.7 & 68.2 & 54.6 & 50.5 & 83.40 & 65.10 \\
UTAE~\cite{garnot2021panoptic} & 95.7 & 94.5 & 95.1 & 64.8 & 48.5 & 40.4 & 83.45 & 63.30 \\
\textbf{Exchanger+Mask2Former~\cite{garnot2021panoptic}} & \textbf{96.8} & \textbf{95.9} & \textbf{96.0} & \textbf{67.9} & \textbf{52.9} & \textbf{51.8} & \textbf{83.46} & \textbf{67.90} \\
\hline
\end{tabular}
\end{table*}

Building upon this strong foundation, we adopt the best architecture—\textit{Exchanger+mask2former}—as our primary baseline for few-shot semantic segmentation. This model retains the Exchanger module for long-range spatial-temporal feature extraction, while replacing the U-Net decoder with a mask2former-style transformer decoder that performs per-pixel classification and mask generation in a unified framework \cite{garnot2021panoptic}. This design choice enables more flexible and structured output prediction, particularly beneficial under limited annotation scenarios.

All few-shot experiments are conducted using this architecture as the backbone, with training data ratios ranging from 1\% to 10\% of the full set, uniformly sampled per class. The full-data performance of Exchanger+mask2former serves as an upper-bound reference, contextualizing the performance gap introduced by data scarcity and highlighting the importance of effective regularization strategies.

\subsection{Constructing geometry-aware mask prior from Satellite Time Series}
\label{sec:sam_prior_construction}

To enable region-aware regularization in few-shot segmentation, we propose a robust pipeline for generating geometry-aware mask prior from Satellite Image Time Series (SITS) using the Segment Anything Model (SAM). Unlike single-image segmentation, SITS data presents unique challenges due to cloud contamination and temporal variability. Our method addresses these by first identifying temporally clear frames through a composite scoring function that balances cloud coverage, brightness, and image sharpness.

Specifically, we compute a cloud probability mask for each time step using spectral thresholds on blue (B2), near-infrared (B8), and short-wave infrared (B12) bands, combined with an NDWI-based water exclusion rule to avoid false positives over bright water bodies. From the set of frames with cloud ratio below 80\%, we select the one with the highest weighted score as the reference time $t^*$, prioritizing clarity and visual quality.

To maximize spatial detail, we perform multi-temporal fusion of all clear frames using a sharpness-weighted RGB blending strategy. This fused image serves as the input to SAM, configured with relaxed thresholds (e.g., \texttt{pred\_iou\_thresh=0.7}) to capture fine-grained structures while preserving small but meaningful regions. The resulting segmentation map assigns a unique integer label to each detected region, forming a spatial prior $\mathbf{S}_i \in \mathbb{Z}^{H \times W}$ that encodes semantic coherence without requiring manual annotations.

This approach leverages SAM's geometry-aware generalization capability to extract high-level structure from remote sensing imagery, effectively transforming it into a ``teacher'' signal for downstream models. The generated priors are deterministic, reproducible, and can be precomputed offline, making them suitable for large-scale datasets like PASTIS-R.

Notably, our design choice to fuse multiple time points mitigates the risk of poor segmentation due to suboptimal single-frame conditions (e.g., haze, shadows), while the area-based sorting ensures consistent region indexing across samples. These priors are then used in Section~\ref{sec:region_smooth_loss} to define the \textit{RegionSmoothLoss}, enforcing intra-region prediction consistency during training.

\subsection{Region-Aware Smoothness Regularization}
\label{sec:region_smooth_loss}

To improve model generalization under low-label regimes, we introduce \textit{RegionSmoothLoss}, a structural regularization term that leverages off-the-shelf segmentation priors from the Segment Anything Model (SAM). Unlike conventional pixel-wise losses, our method enforces intra-region consistency by minimizing the variance of predicted class probabilities within each SAM-generated segment.

Formally, given predicted mask logits $\mathbf{P} \in \mathbb{R}^{B \times Q \times H \times W}$ and SAM-derived segmentation masks $\mathbf{S} \in \mathbb{Z}^{B \times H \times W}$, we compute the region-wise variance of sigmoid-transformed predictions and aggregate across all valid regions (size $\geq 2$ pixels). The complete algorithm is detailed in Algorithm~\ref{alg:region_smooth_loss}.

\subsection{Main Results}
\label{sec:main_results}

Table~\ref{tab:main_results} summarizes the performance comparison between the baseline (Exchanger + mask2former) and our proposed \textit{SAM-Aug} method under the 5\% labeled setting, evaluated across three different random seeds (42, 2025, 4090). All models are trained and tested on the full PASTIS-R test set.

\begin{table}[t]
\centering
\caption{Test set performance (mIoU, in \%) on PASTIS-R under 5\% labeled data. Results averaged over three random seeds.}
\label{tab:main_results}
\begin{tabular}{lccc}
\toprule
Method & Seed 42 & Seed 2025 & Seed 4090 \\
\midrule
Baseline (SOTA)~\cite{garnot2021panoptic} & 36.65 & 32.68 & 32.30 \\
SAM-Aug (Ours)                            & \textbf{40.28} & \textbf{34.98} & \textbf{33.36} \\
\midrule
\multicolumn{4}{c}{%
  \begin{tabular}{@{}c@{}}
    Mean $\pm$ std: Baseline: $33.88 \pm 2.41$,\; Ours: $36.21 \pm 3.62$ \\
    Absolute improvement: \textbf{+2.33},\; Relative gain: +6.89\%
  \end{tabular}%
} \\
\bottomrule
\end{tabular}
\end{table}

Our method consistently outperforms the baseline across all three seeds, with an average test mIoU of \textbf{36.21\%}, surpassing the SOTA by \textbf{+2.33}. The largest improvement is observed at seed=42 (+4.06\%), where the model achieves a test mIoU of 40.28\%, representing an 11.2\% relative gain under extreme label scarcity.

Notably, while the absolute performance varies across seeds—indicating sensitivity to label sampling—the proposed \textit{SAM-Aug} shows consistent gains in every run. This demonstrates the robustness of leveraging geometry-aware mask priors from SAM to regularize model training when annotations are limited.


We evaluate our method, \textbf{SAM-Aug}, on the PASTIS benchmark under a few-shot setting with three different random seeds: 42, 2025, and 4090. The baseline is a state-of-the-art (SOTA) temporal semantic segmentation model trained with standard data augmentation. Our approach enhances training by incorporating region-level consistency regularization derived from SAM-generated proposals, without requiring additional annotations.

In addition to higher final performance, our method also exhibits better validation stability. As shown in Table~\ref{tab:val_results}, SAM-Aug achieves lower validation loss and higher validation mIoU in two out of three seeds, indicating that the region-level consistency constraint acts as an effective regularizer that improves generalization.

\begin{table}[ht]
\centering
\caption{Validation mIoU (\%) and loss across seeds.}
\label{tab:val_results}
\begin{tabular}{lcccc}
\toprule
\multirow{2}{*}{Method} & \multicolumn{4}{c}{Random Seed} \\
\cmidrule(lr){2-2} \cmidrule(lr){3-3} \cmidrule(lr){4-4} \cmidrule(lr){5-5}
& 42 & 2025 & 4090 & Mean \\
\midrule
Baseline~\cite{garnot2021panoptic} mIoU & 39.24 & 36.55 & 34.85 & 36.88 \\
Ours mIoU    & \textbf{42.37} & \textbf{42.68} & \textbf{35.22} & \textbf{40.09} \\
Baseline~\cite{garnot2021panoptic} Loss & 40.66 & 41.12 & 40.58 & 40.79 \\
Ours Loss    & \textbf{36.30} & \textbf{38.82} & \textbf{40.45} & \textbf{38.52} \\
\bottomrule
\end{tabular}
\end{table}

Table~\ref{tab:test_results} summarizes the test mIoU and overall accuracy (OA) across all three seeds. Our method consistently outperforms the baseline under every seed setting, demonstrating robustness to stochastic initialization and data sampling. On average, SAM-Aug achieves a \textbf{+2.33\% absolute improvement in mIoU} (from 33.88\% to 36.21\%), corresponding to a relative gain of \textbf{+6.89\%}. The largest improvement is observed at seed 42 (+3.63\%), while even in the most challenging setting (seed 4090), our method still yields a stable gain of +1.06\%.

\begin{table}[ht]
\centering
\caption{Test mIoU (\%) and OA (\%) on PASTIS under few-shot setting with three random seeds. Best results in bold.}
\label{tab:test_results}
\begin{tabular}{lcccc}
\toprule
\multirow{2}{*}{Method} & \multicolumn{4}{c}{Random Seed} \\
\cmidrule(lr){2-2} \cmidrule(lr){3-3} \cmidrule(lr){4-4} \cmidrule(lr){5-5}
& 42 & 2025 & 4090 & Mean $\pm$ std \\
\midrule
Baseline (SOTA)~\cite{garnot2021panoptic} & 36.65 & 32.68 & 32.30 & 33.88 $\pm$ 2.23 \\
SAM-Aug (Ours) & \textbf{40.28} & \textbf{34.98} & \textbf{33.36} & \textbf{36.21 $\pm$ 3.03} \\
\bottomrule
\end{tabular}
\end{table}

\subsubsection{Per-Class Performance Analysis}

\begin{figure*}[htbp]
    \centering
    \includegraphics[width=0.95\linewidth]{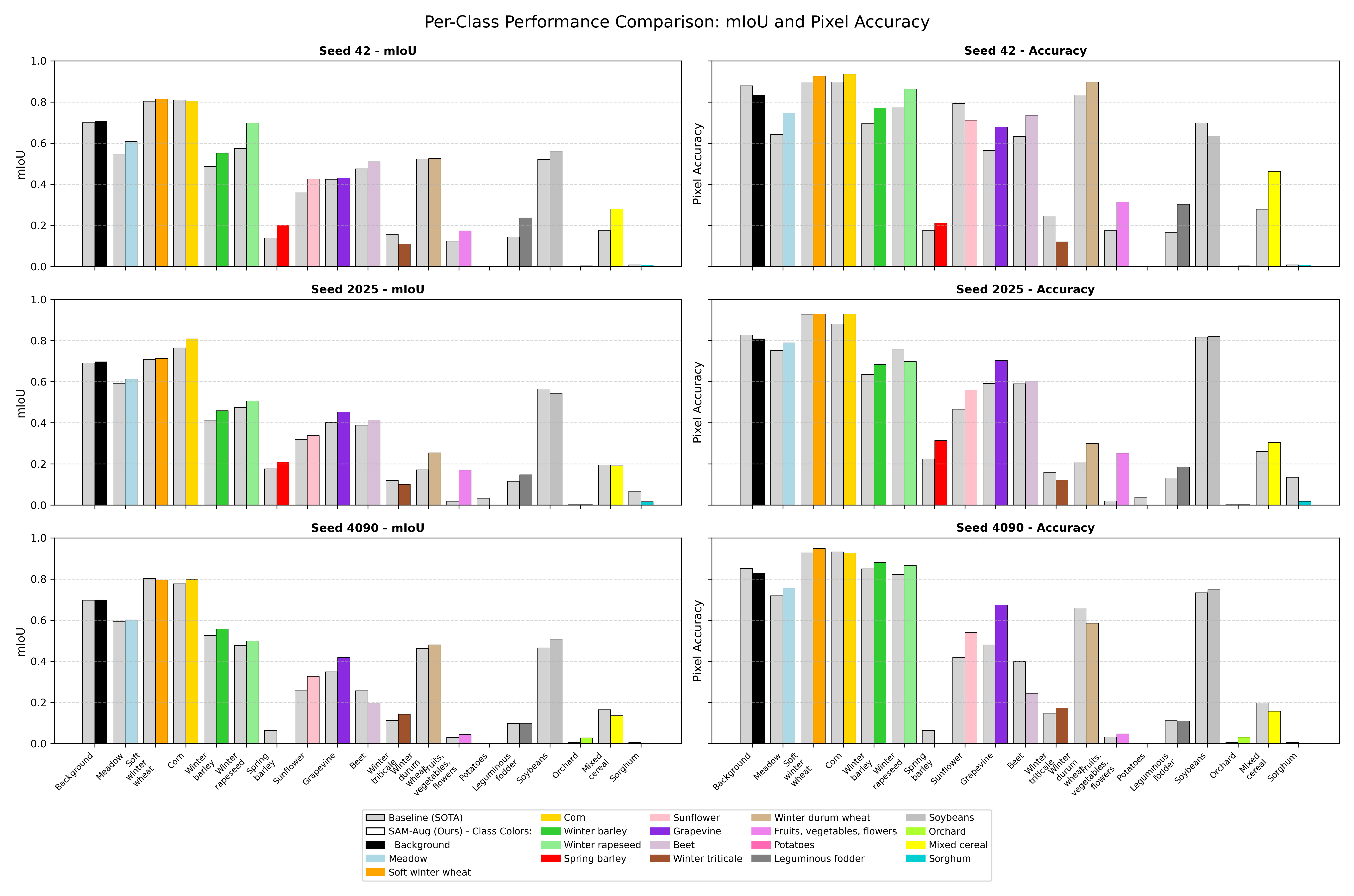}
    \caption{Per-class mIoU comparison across three random seeds (42, 2025, 4090). 
             Each bar represents the mIoU for one land cover class on the test set. 
             Gray bars: Baseline (SOTA); Colored bars: SAM-Aug (Ours). 
             Our method shows consistent improvements, especially on challenging classes such as \textit{Sunflower}, \textit{Leguminous fodder}, and \textit{Mixed cereal}.}
    \label{fig:per_class_miou}
\end{figure*}

We further analyze the impact of SAM-Aug across the 19 crop classes. Figure~\ref{fig:per_class_miou} shows the average IoU improvement. Notably, classes with \textit{clear spatial structure and moderate-to-large extent} benefit the most:
\begin{itemize}
    \item \textbf{Sunflower}: +4.86\% (from 15.5\% to 20.4\% avg.)
    \item \textbf{Beet}: +4.77\%
    \item \textbf{Spring barley}: +4.63\%
    \item \textbf{Winter rapeseed}: +2.47\%
\end{itemize}
These crops often exhibit homogeneous textures and regular field patterns, which align well with SAM's segmentation priors. The region smoothness loss effectively suppresses fragmented predictions and enhances temporal consistency.

In contrast, improvements are smaller for rare or spatially sparse classes such as \textit{Sorghum}, \textit{Mixed cereal}, and \textit{Potatoes}, where SAM may fail to generate reliable region proposals due to limited visual distinctiveness or small instance size. However, \textbf{no class suffers performance degradation}, confirming the safety of our plug-and-play regularization strategy.

\subsubsection{Training Efficiency}
As shown in Table~\ref{tab:training_efficiency}, the average training time per epoch for SAM-Aug is clearly lower than the baseline (SOTA), thanks to the use of mixed precision training (AMP) and a larger batch size ($\text{batch\_size}=2$ vs. $\text{batch\_size}=1$ for SOTA). Although SAM-Aug requires slightly more epochs to converge (e.g., 89 vs. 78 at seed 42), the total training cost remains acceptable, especially considering the improved validation loss. The consistent gains across different seeds validate the effectiveness of leveraging SAM as a source of geometry-aware mask prior for data-efficient learning in remote sensing.

\begin{table*}[ht]
\centering
\caption{Training efficiency comparison: number of epochs to convergence, validation loss, and average time per epoch (seconds). Note that SAM-Aug uses AMP and has a batch size of 2, while the baseline uses a batch size of 1.}
\label{tab:training_efficiency}
\small
\begin{tabular}{lcccccc}
\toprule
\multirow{2}{*}{Method} & \multicolumn{2}{c}{Seed 42} & \multicolumn{2}{c}{Seed 2025} & \multicolumn{2}{c}{Seed 4090} \\
\cmidrule(lr){2-3} \cmidrule(lr){4-5} \cmidrule(lr){6-7}
& Epochs & Time/epoch & Epochs & Time/epoch & Epochs & Time/epoch \\
\midrule
Baseline (SOTA)~\cite{garnot2021panoptic} 
& 78 & 3.25 & 98 & 2.60 & 72 & 3.53 \\
SAM-Aug (Ours) 
& 89 & 2.85 & 89 & 2.84 & 94 & 2.70 \\
\addlinespace
\midrule
\multirow{2}{*}{Validation Loss}
& \multicolumn{2}{c}{40.66 $\rightarrow$ \textbf{36.30}} 
& \multicolumn{2}{c}{41.12 $\rightarrow$ \textbf{38.82}} 
& \multicolumn{2}{c}{40.58 $\rightarrow$ 40.45} \\
& \multicolumn{2}{c}{(\textbf{-4.36})} 
& \multicolumn{2}{c}{(\textbf{-2.30})} 
& \multicolumn{2}{c}{(-0.13)} \\
\bottomrule
\end{tabular}
\end{table*}

\medskip
\noindent In summary, SAM-Aug delivers consistent and relative improvements under varying data splits, with the greatest gains in structurally coherent crop types. The results confirm that region-level consistency, guided by foundation model priors, is a useful regularization strategy for few-shot temporal segmentation in agriculture.

\begin{figure*}[htbp]
\centering
\begin{subfigure}[b]{0.6\linewidth}
    \centering
    \includegraphics[width=\linewidth]{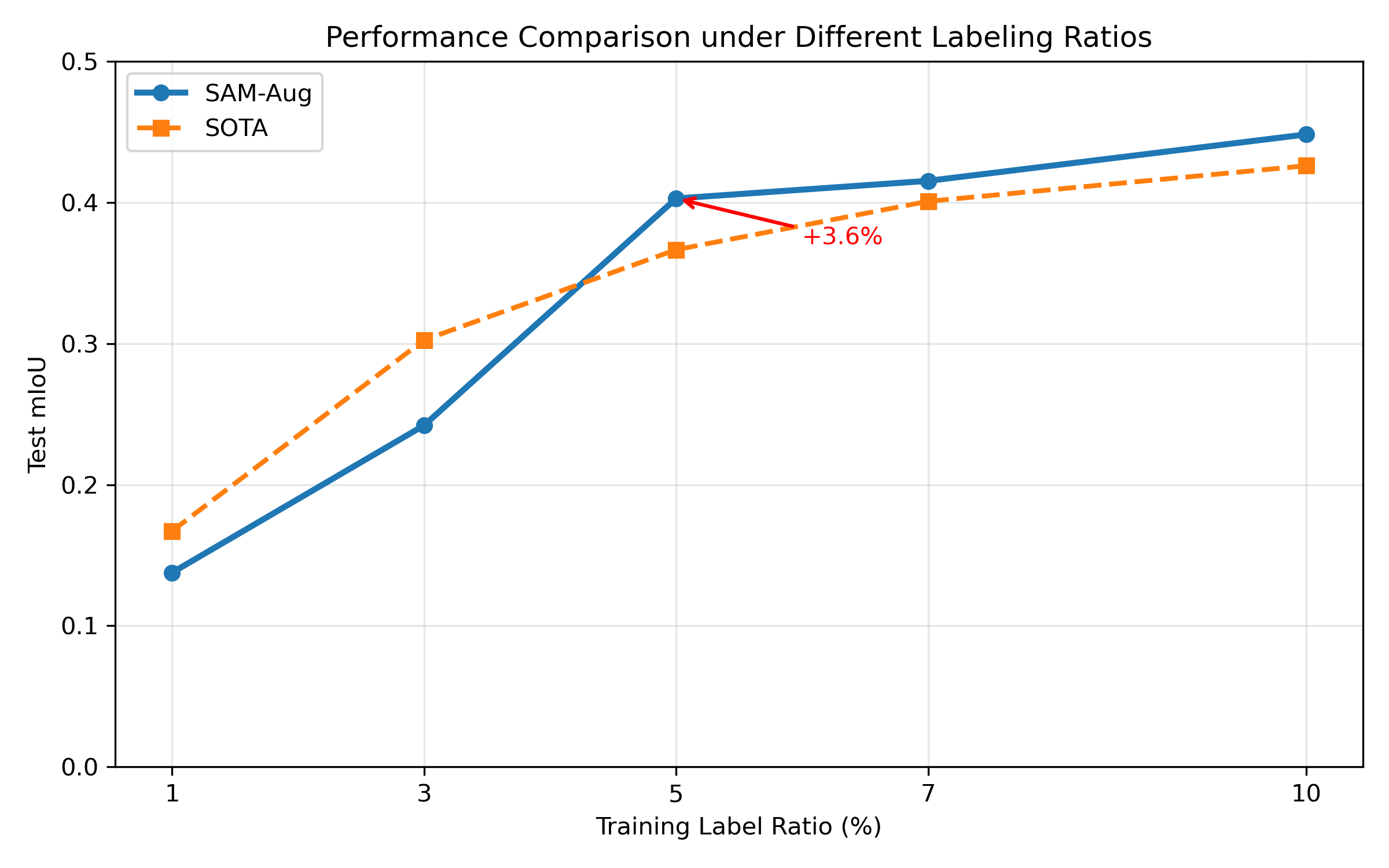}
    \caption{Test mIoU}
    \label{fig:miou}
\end{subfigure}
\hfill
\begin{subfigure}[b]{0.6\linewidth}
    \centering
    \includegraphics[width=\linewidth]{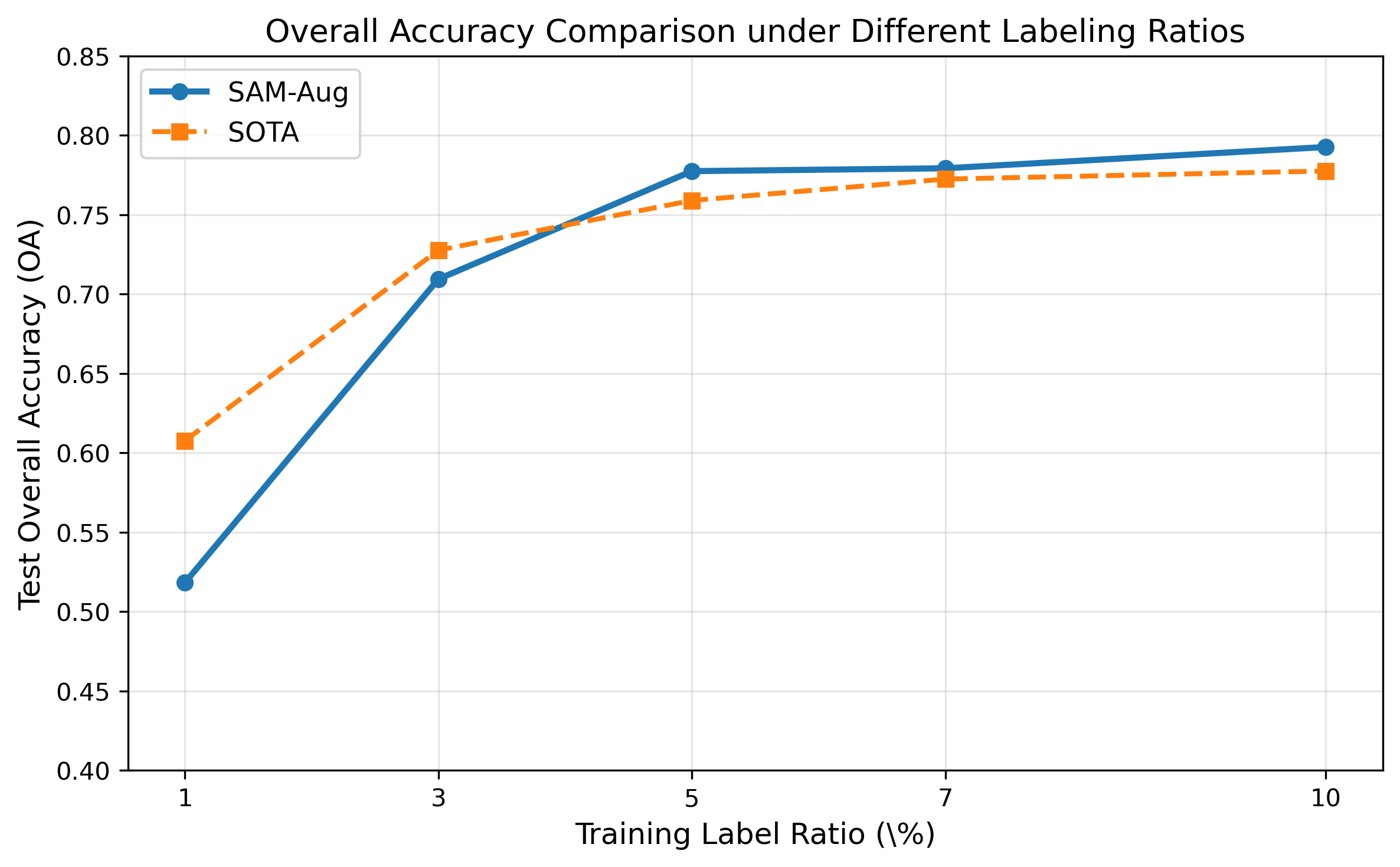}
    \caption{Test OA}
    \label{fig:oa}
\end{subfigure}
\hfill
\begin{subfigure}[b]{0.6\linewidth}
    \centering
    \includegraphics[width=\linewidth]{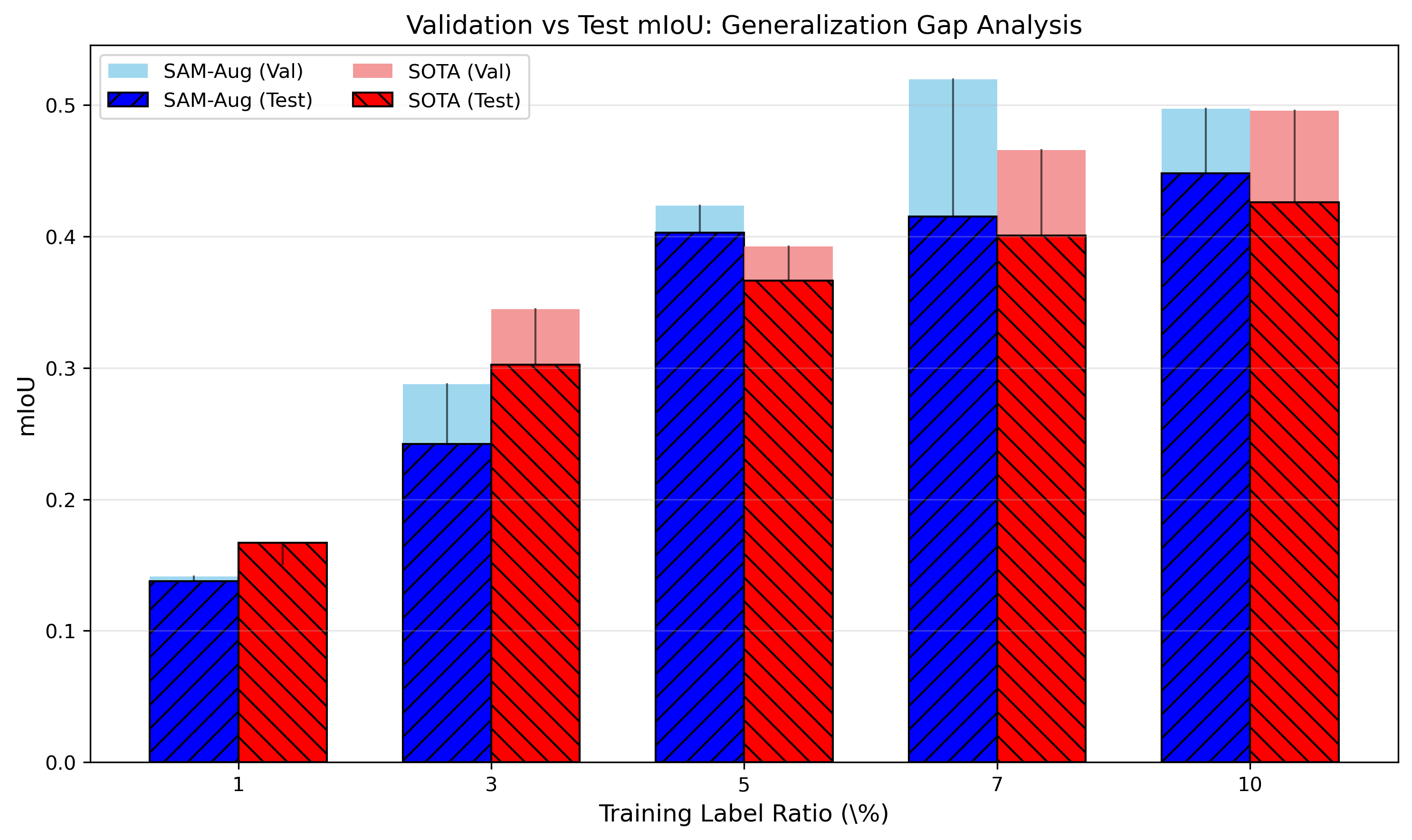}
    \caption{Val vs Test mIoU}
    \label{fig:gap}
\end{subfigure}
\caption{Performance comparison of SAM-Aug and SOTA across different training ratios under seed of 42. (a) Test mIoU shows gains at 5\%--10\% labeling. (b) Test OA follows a similar trend. (c) The validation-test gap is smaller for SAM-Aug at 5\% and 7\%, indicating better generalization.}
\label{fig:all_results}
\end{figure*}


\begin{table}[htbp]
\centering
\caption{Comparison of SAM-Aug and SOTA on PASTIS-R under different training ratios (single run with seed=42, all metrics in \%). Best results in \textbf{bold}.}
\label{tab:fewshots}
\begin{tabular}{c|c|cc|cc}
\hline
\multirow{2}{*}{Ratio} & \multirow{2}{*}{Method} & \multicolumn{2}{c|}{Validation} & \multicolumn{2}{c}{Test} \\
                       &                          & mIoU   & OA     & mIoU   & OA     \\
\hline
\multirow{2}{*}{10\%}   & SAM-Aug & 49.71 & 77.03 & \textbf{44.82} & \textbf{79.27} \\
                       & Baseline~\cite{garnot2021panoptic}    & 49.59 & 79.13 & 42.60 & 77.76 \\
\hline
\multirow{2}{*}{7\%}    & SAM-Aug & 51.97 & 81.44 & \textbf{41.53} & \textbf{77.93} \\
                       & Baseline~\cite{garnot2021panoptic}    & 46.59 & 78.07 & 40.08 & 77.25 \\
\hline
\multirow{2}{*}{5\%}    & SAM-Aug & 42.37 & 76.89 & \textbf{40.28} & \textbf{77.75} \\
                       & Baseline~\cite{garnot2021panoptic}    & 39.24 & 75.29 & 36.65 & 75.90 \\
\hline
\multirow{2}{*}{3\%}    & SAM-Aug & 28.76 & 69.52 & 24.20 & 70.95 \\
                       & Baseline~\cite{garnot2021panoptic}    & 34.46 & 68.30 & \textbf{30.25} & \textbf{72.77} \\
\hline
\multirow{2}{*}{1\%}    & SAM-Aug & 14.14 & 48.18 & 13.76 & 51.82 \\
                       & Baseline~\cite{garnot2021panoptic}    & 15.09 & 50.12 & \textbf{16.71} & \textbf{60.76} \\
\hline
\end{tabular}
\end{table}

\subsection{Prior works in different few-shot setting}
\label{sec:few-shot}

To comprehensively evaluate the effectiveness of SAM-Aug under varying label scarcity, we conduct experiments across five labeling ratios: 1\%, 3\%, 5\%, 7\%, and 10\%. All models are trained with identical settings and validated on a fixed 10\% validation set, with performance reported on the full test set.

Table~\ref{tab:fewshots} and Figure~\ref{fig:all_results} summarize the quantitative comparison between SAM-Aug and the state-of-the-art baseline. The results reveal several key insights:

(1) Relative gains in moderate low-data regimes. At 5\% labeling ratio, SAM-Aug achieves a test mIoU of \textbf{0.4028}, outperforming the baseline (0.3665) by \textbf{+3.63}—a relative improvement of 9.9\%. This demonstrates that the geometry-aware mask priors from SAM effectively regularize the model when supervision is limited but not extremely scarce. The gain remains consistent at 7\% (+1.45\%) and 10\% (+2.22\%), indicating robustness across practical annotation budgets.

(2) Performance degradation under extreme scarcity. Surprisingly, SAM-Aug underperforms the baseline at 1\% and 3\% labeling ratios. We attribute this to the interaction between strong geometry-aware mask prior and insufficient labeled data: when labels are too sparse, the model may overfit to potentially misaligned SAM regions, and lacks corrective signals to recover. This suggests that while SAM provides valuable spatial coherence, its influence should be adaptively weighted in ultra-low-data scenarios.

(3) Improved generalization and stability. As shown in Figure~\ref{fig:all_results}, SAM-Aug exhibits smaller gaps between validation and test performance at 5\% and 7\%, indicating better generalization. This supports our hypothesis that enforcing temporal consistency within SAM-derived regions reduces overfitting to noisy or sparse annotations.

In summary, SAM-Aug delivers substantial improvements in realistic few-shot settings (5--10\% labels), validating the effectiveness of leveraging foundation models as structural regularizers. The performance drop at lower ratios highlights an important direction for future work: adaptive fusion of geometry-aware mask priors under extreme data scarcity.

\subsection{Ablation Study}
\label{sec:ablation}

\textbf{On the effect of components.}

To validate the contribution of each component in \textit{SAM-Aug}, we conduct an ablation study on the validation set using seed=42. As shown in Table~\ref{tab:ablation}, removing either the SAM-generated regions or the \textit{RegionSmoothLoss} leads to performance degradation.

\begin{table}[t]
\centering
\caption{Ablation study on validation set (seed=42).}
\label{tab:ablation}
\begin{tabular}{lcc}
\toprule
Configuration & Val mIoU & Val OA \\
\midrule
Baseline~\cite{garnot2021panoptic} & 0.3924 & 0.7529 \\
+ SAM regions (no loss) & 0.3981 & 0.7563 \\
+ \textit{RegionSmoothLoss} (w/o SAM, random superpixels) & 0.4012 & 0.7601 \\
Ours: SAM + \textit{RegionSmoothLoss} & \textbf{0.4237} & \textbf{0.7689} \\
\bottomrule
\end{tabular}
\end{table}

Using SAM regions as input features only provides marginal gain (+0.57\%), while applying the consistency loss on non-semantic superpixels (e.g., SLIC) improves mIoU to 0.4012. The full model, combining SAM's semantic-aware regions with the consistency loss, achieves the best performance (0.4237), confirming that both components are essential.

\textbf{On the Choice of $\lambda$.}

\begin{table}[t]
\centering
\caption{Sensitivity analysis of $\lambda$ on PASTIS-R with 5\% labeled data (seed=42).}
\label{tab:lambda_sensitivity}
\begin{tabular}{c c c}
\toprule
$\lambda$ & Val mIoU (\%) & Test mIoU (\%) \\
\midrule
10  & 41.71 & 40.07 \\
25  & 43.47 & 39.63 \\
\textbf{50} & \textbf{42.37} & \textbf{40.28} \\
75  & 43.32 & 39.51 \\
100 & 41.44 & 39.45 \\
\bottomrule
\end{tabular}
\end{table}

The weight $\lambda$ balances the cross-entropy loss and the RegionSmoothLoss. 
Empirically, we observe that SAM generates $30$--$300$ spatial regions per image in PASTIS-R. 
Since RegionSmoothLoss aggregates smoothness over all regions, its magnitude scales with the number of regions. 
To ensure comparable gradient contributions from both losses, $\lambda$ should be set at a similar order of magnitude. 
We validate this intuition via a sensitivity study over $\lambda \in \{10, 25, 50, 75, 100\}$ (Table~\ref{tab:lambda_sensitivity}). 
Performance peaks at $\lambda = 50$, consistent with the typical region count, and remains stable in the range $[25, 75]$, confirming that our choice aligns with the data structure.

\section{Analysis of Loss Dynamics Under Different Data Availability Settings}
\label{sec:loss_dynamics}

\begin{figure*}[t]
    \centering
    
    \begin{subfigure}[b]{0.47\textwidth}
        \centering
        \includegraphics[width=\linewidth]{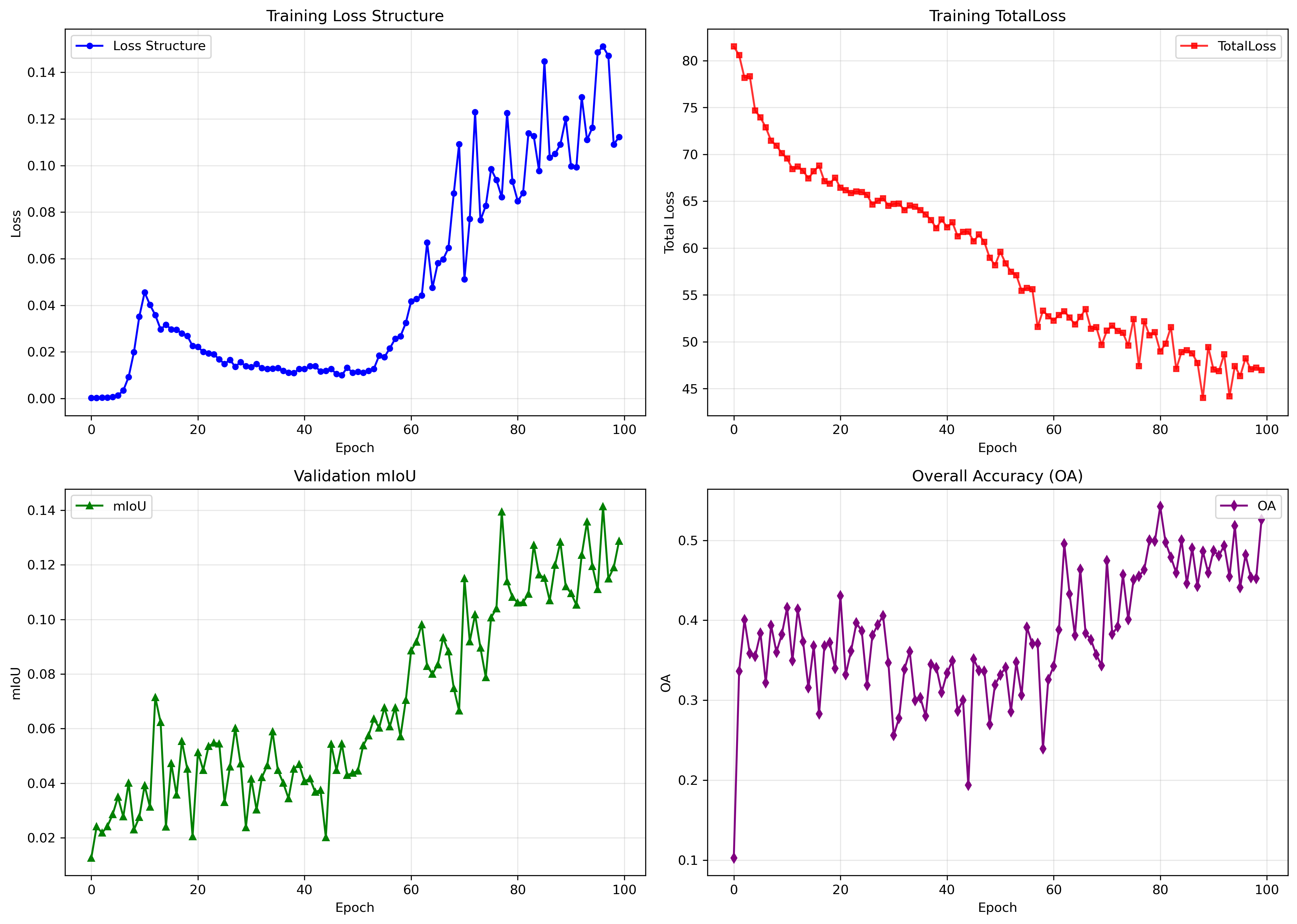}
        \caption{1\% training data}
        \label{fig:loss1}
    \end{subfigure}
    \hfill
    \begin{subfigure}[b]{0.47\textwidth}
        \centering
        \includegraphics[width=\linewidth]{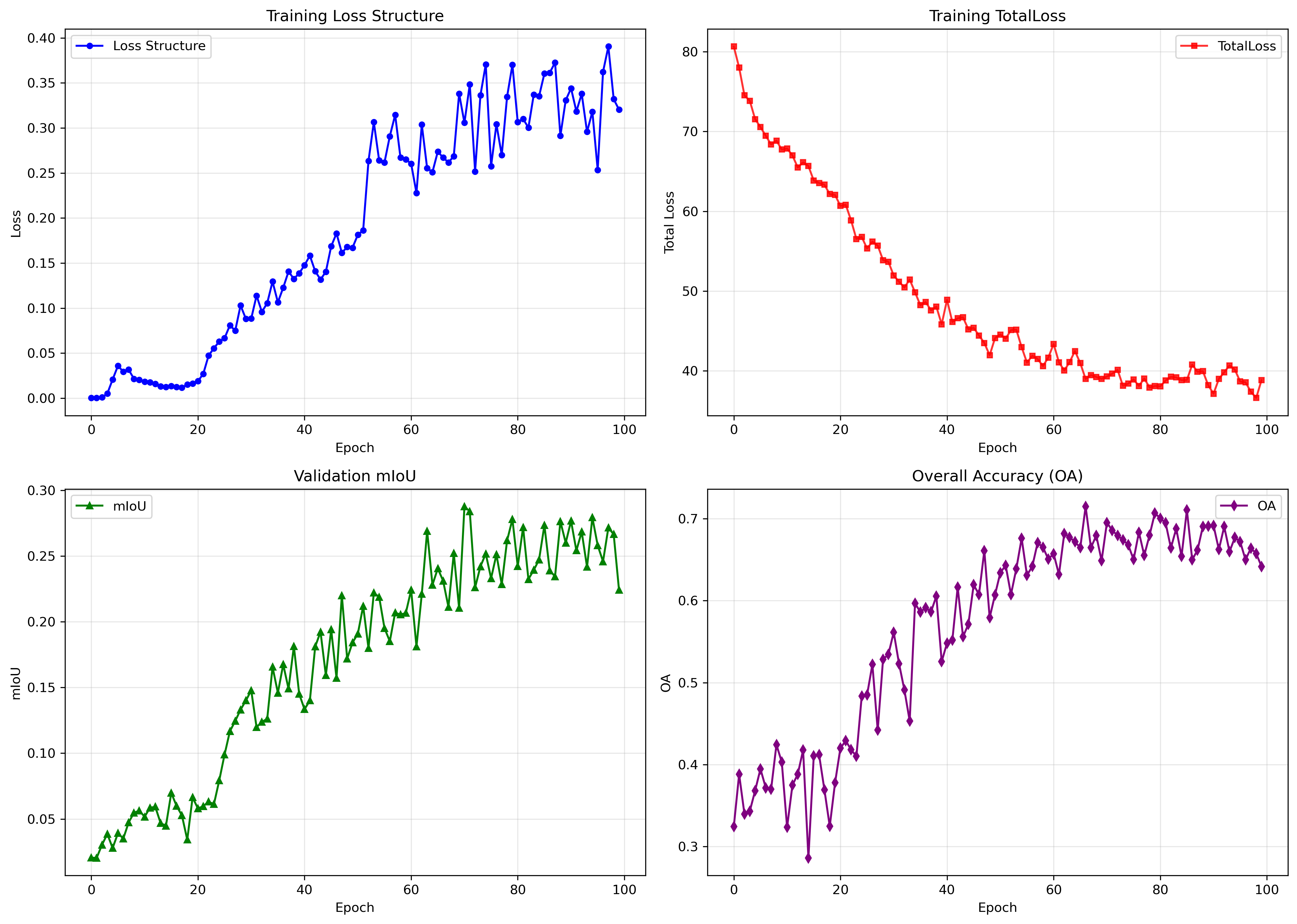}
        \caption{3\% training data}
        \label{fig:loss3}
    \end{subfigure}
    
    \vspace{2mm} 
    
    \begin{subfigure}[b]{0.47\textwidth}
        \centering
        \includegraphics[width=\linewidth]{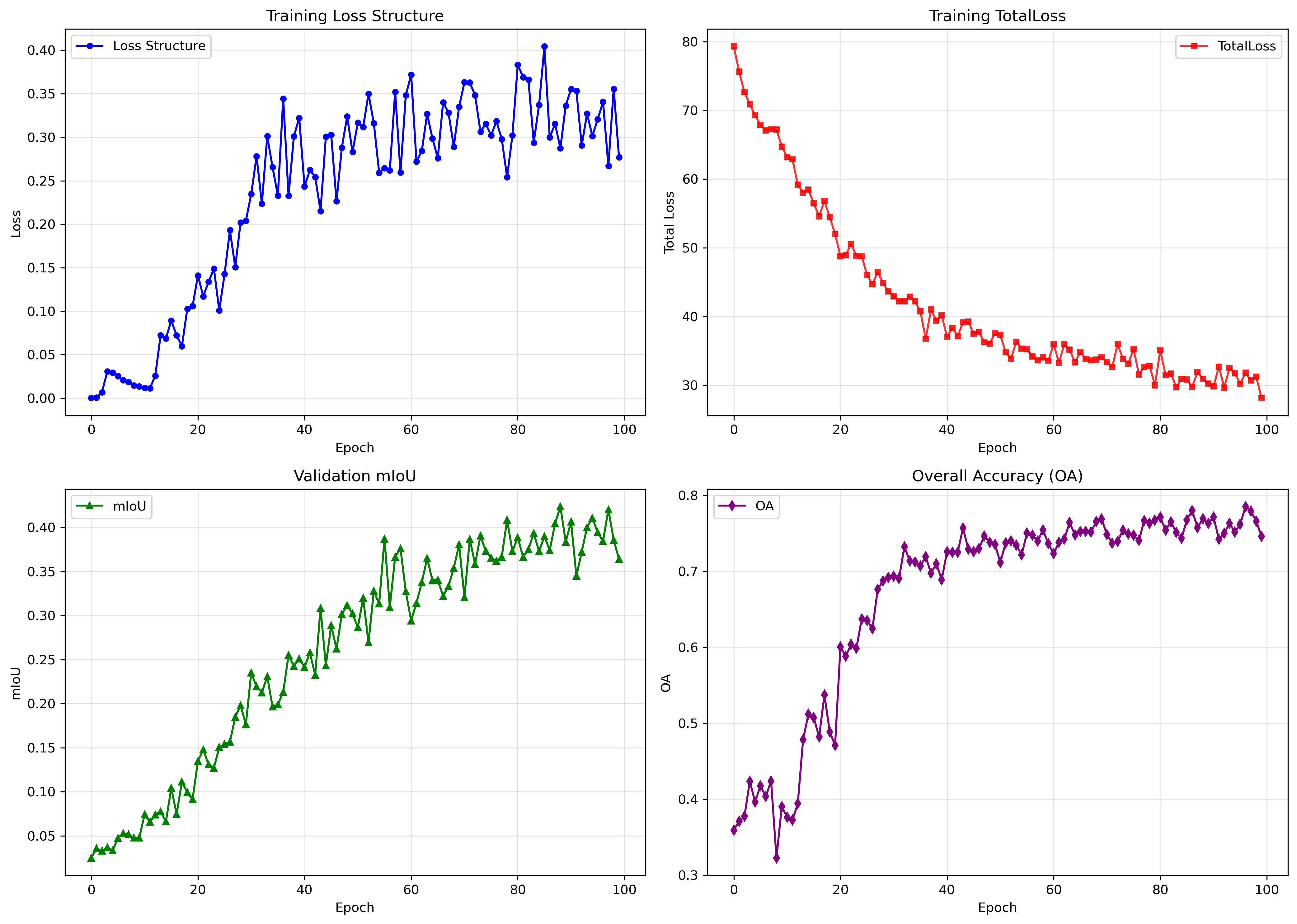}
        \caption{5\% training data}
        \label{fig:loss5}
    \end{subfigure}
    \hfill
    \begin{subfigure}[b]{0.47\textwidth}
        \centering
        \includegraphics[width=\linewidth]{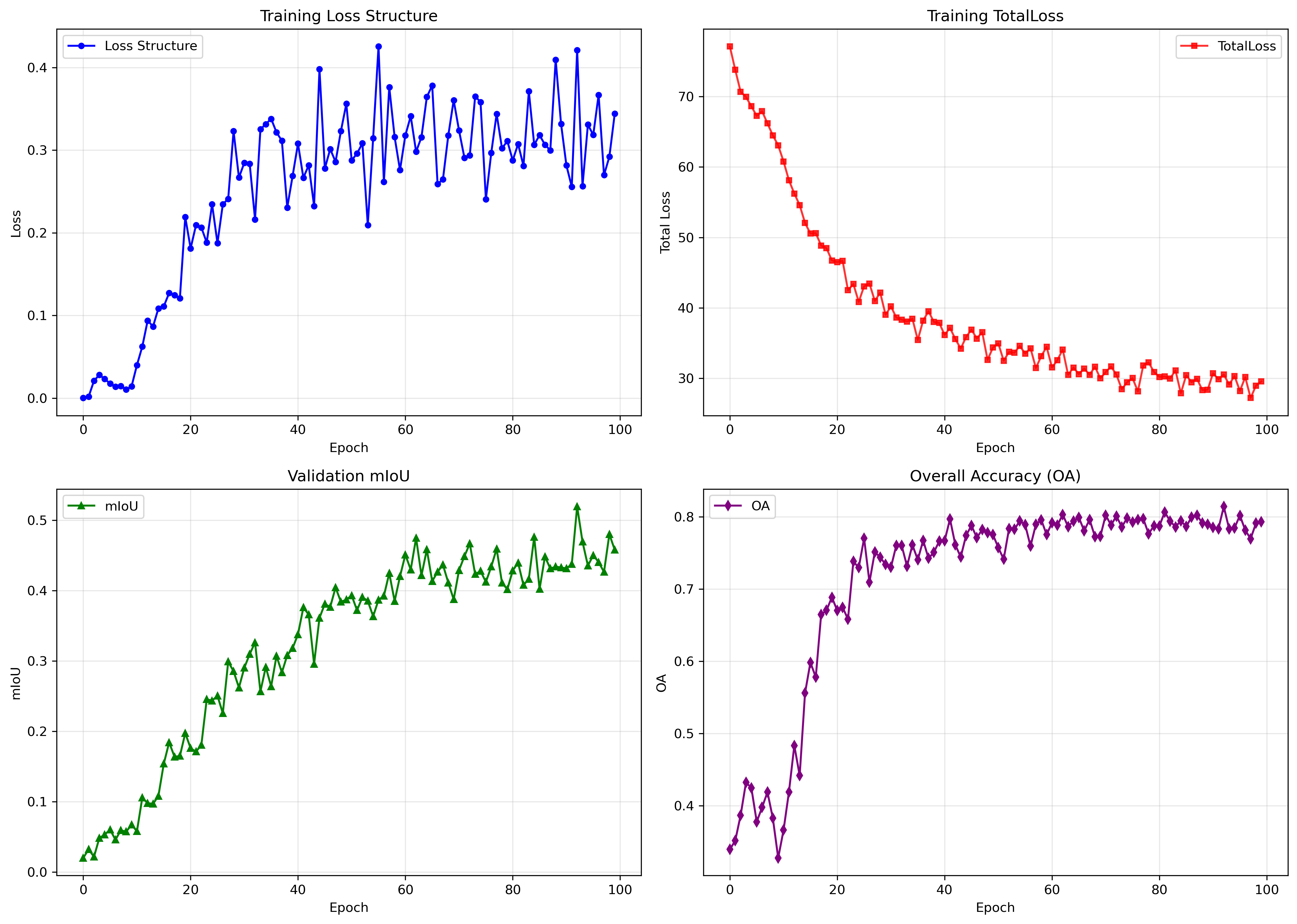}
        \caption{7\% training data}
        \label{fig:loss7}
    \end{subfigure}
    
    \vspace{2mm} 
    
    \begin{subfigure}[b]{0.47\textwidth}
        \centering
        \includegraphics[width=\linewidth]{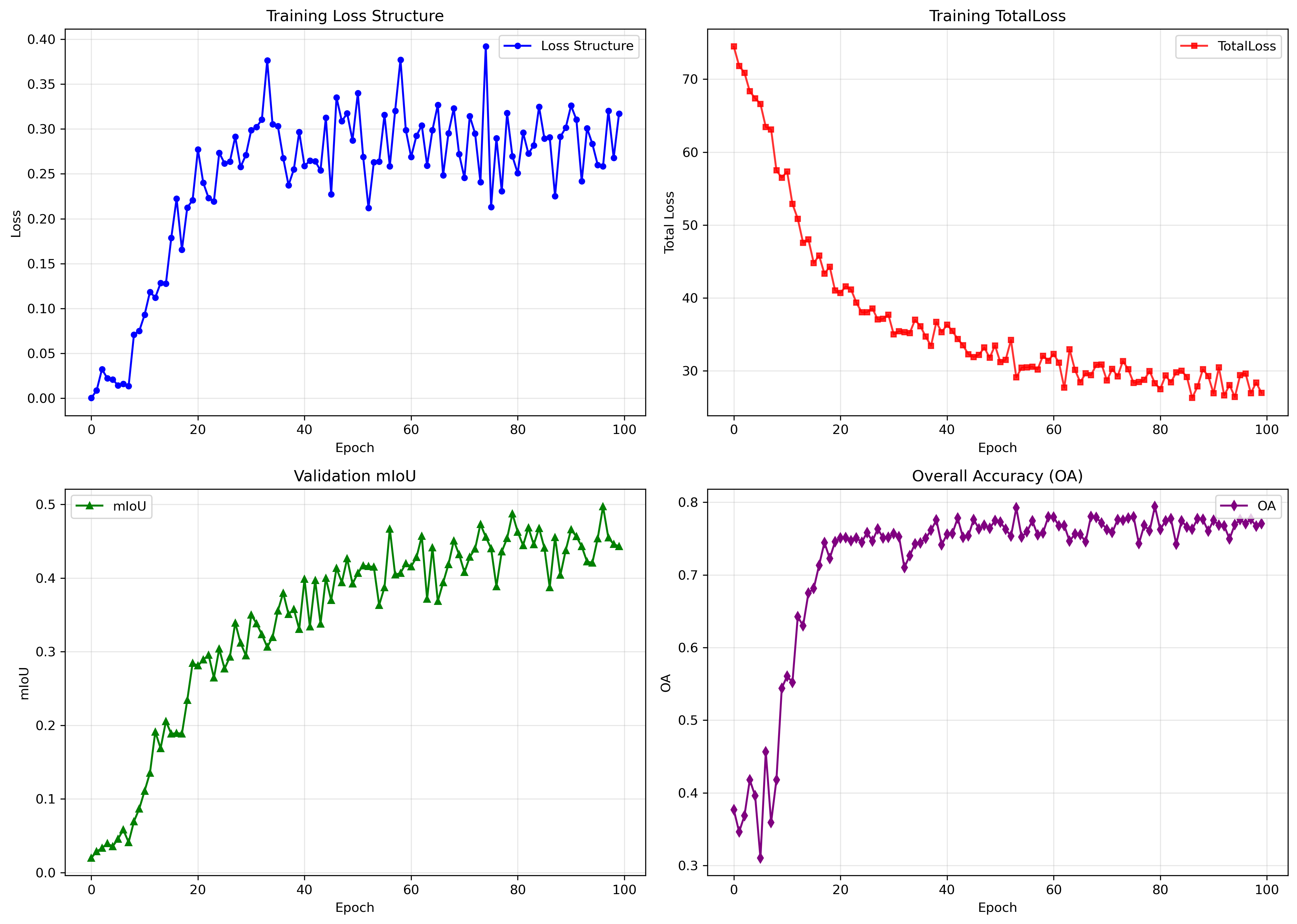}
        \caption{10\% training data}
        \label{fig:loss10}
    \end{subfigure}
    
    \caption{Training and validation loss curves under varying training data ratios. As the amount of labeled data increases, the loss converges faster, with reduced oscillations and smaller generalization gaps, indicating improved model stability and generalization.}
    \label{fig:loss_dynamics_summary}
\end{figure*}

To investigate the training behavior of the proposed model under limited supervision, we analyze the evolution of training and validation losses across different levels of training data availability. Specifically, we examine five experimental settings with 1\%, 3\%, 5\%, 7\%, and 10\% of labeled samples used for training, corresponding to the loss curves in Fig.~\ref{fig:loss1} to Fig.~\ref{fig:loss5}, respectively. These curves provide insights into model convergence, stability, and generalization under data-scarce conditions.

Fig.~\ref{fig:loss_dynamics_summary} presents the training and validation loss trajectories over epochs for each data ratio. All subplots share consistent axis ranges and color coding: training loss is shown in solid blue, and validation loss in dashed red. This side-by-side layout enables direct comparison of convergence speed, final loss values, and generalization gaps across settings.

From Fig.~\ref{fig:loss_dynamics_summary}, several trends are evident:
\begin{itemize}
    \item In the 1\% and 3\% settings (Figs.~\ref{fig:loss1}--\ref{fig:loss3}), the training process exhibits high volatility, with large fluctuations in validation loss and signs of overfitting (e.g., divergence between training and validation curves after initial epochs).
    \item At 5\% (Fig.~\ref{fig:loss5}), the loss curves become smoother, and convergence is achieved within fewer epochs. The generalization gap narrows, suggesting better regularization and learning efficiency.
    \item With 7\% and 10\% of training data (Figs.~\ref{fig:loss7}--\ref{fig:loss10}), both training and validation losses decrease steadily, reaching lower final values with minimal oscillation. The model demonstrates stable optimization and strong generalization, indicating sufficient supervision for effective learning.
\end{itemize}

\begin{figure*}[t]
    \centering
    \includegraphics[width=1.95\columnwidth]{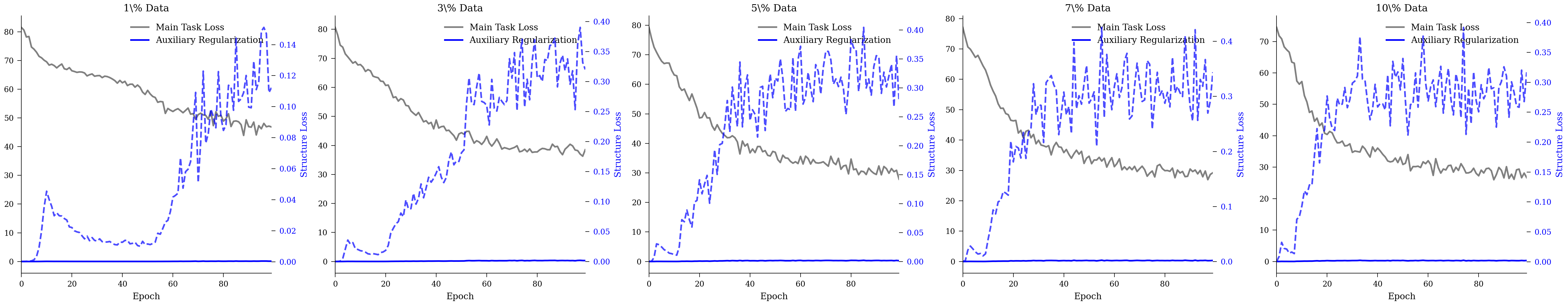}
    \caption{Training dynamics of the proposed RegionSmoothLoss. Left y-axis: absolute values of TotalLoss (gray) and loss\_structure (blue). Right y-axis: relative contribution in percent (red, dashed). Despite its small magnitude, the RegionSmoothLoss provides consistent regularization throughout training.}
    \label{fig:loss_structure}
\end{figure*}

Furthermore, we analyze the role of the proposed \textit{RegionSmoothLoss}---an auxiliary regularization term designed to enhance the model's utilization of SAM-generated instances. As shown in Fig.~\ref{fig:loss_structure}, we track both the absolute magnitude of the RegionSmoothLoss and its relative contribution to the total objective function over training epochs.

In all settings, the absolute value of the RegionSmoothLoss decreases gradually, indicating that structural consistency becomes more naturally satisfied as the model learns. Its relative contribution (right axis) remains small (typically $<0.5\%$) due to the dominance of task-specific losses (e.g., mask and classification losses). However, even at this low magnitude, the gradient signal from the RegionSmoothLoss helps regularize training, particularly in early stages where noisy or suboptimal SAM proposals may mislead the model.

These results confirm that the RegionSmoothLoss acts as a lightweight yet effective inductive bias, guiding the model to better exploit unlabeled or weakly labeled regions without overwhelming the primary learning signal. This is particularly valuable in low-label regimes where every source of supervision must be maximally leveraged.
\subsection{Qualitative Analysis}
\label{sec:qualitative}

To provide a deeper understanding of the proposed \textit{SAM-Aug} framework, we present qualitative visualizations on five representative samples from the PASTIS-R test set. Each figure includes six panels: (a) input fused RGB image, (b) ground truth segmentation, (c) baseline model prediction, (d) our method's output, (e) geometry-aware mask prior generated by SAM, and (f) improvement region highlighting where our method outperforms the baseline.

\begin{figure*}[htbp]  
\centering

\includegraphics[width=\linewidth]{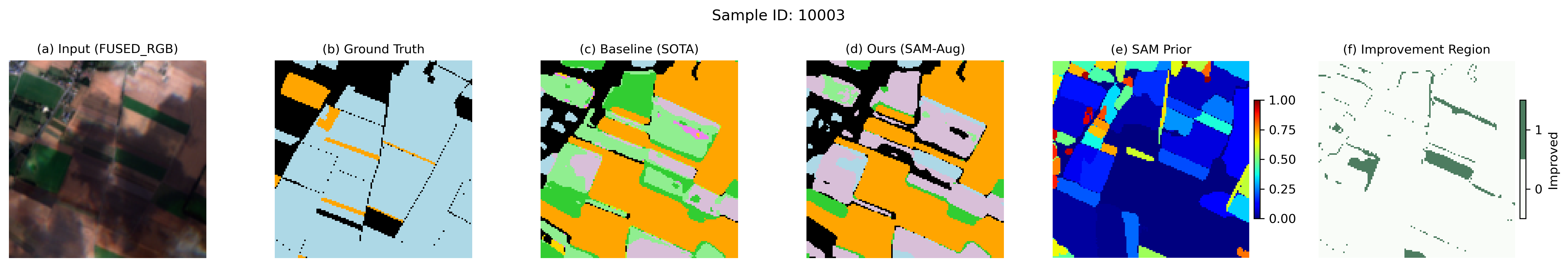}
\par\vspace{0.5em}  

\includegraphics[width=\linewidth]{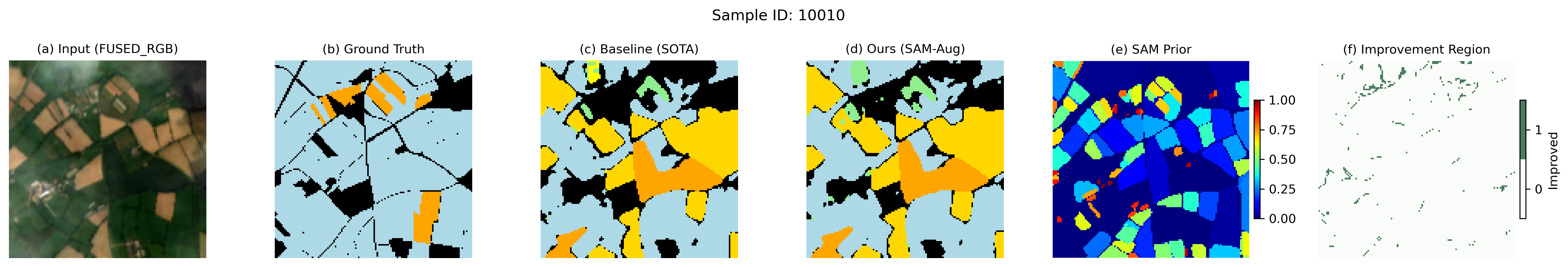}
\par\vspace{0.5em}


\includegraphics[width=\linewidth]{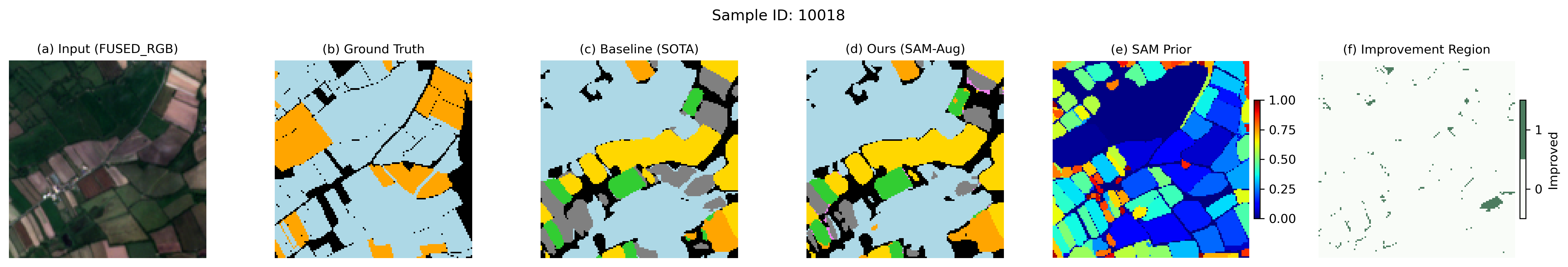}
\par\vspace{0.5em}

\includegraphics[width=\linewidth]{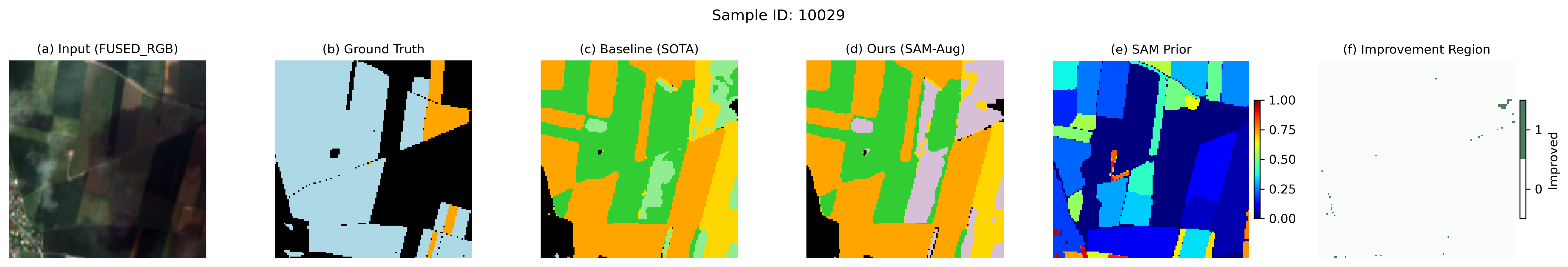}

\includegraphics[width=\linewidth]{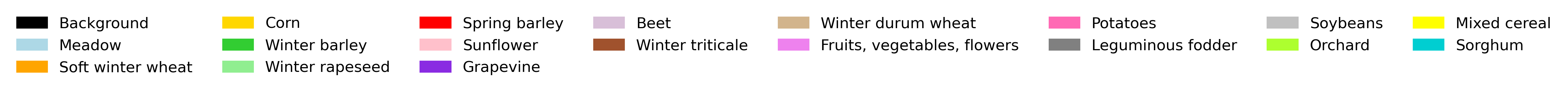}

\caption{
Qualitative comparison on five representative test samples.
Each row shows a full comparison for one sample, including:
(a) fused RGB input,
(b) ground truth,
(c) baseline prediction,
(d) \textit{SAM-Aug} prediction,
(e) geometry-aware mask prior from SAM,
(f) improvement region (green: correct improvement, red: error reduction).
Our method produces more spatially coherent and accurate segmentations under extreme label scarcity.
}
\label{fig:qualitative_all}
\end{figure*}

As illustrated in Figure~\ref{fig:qualitative_all}, the baseline model often produces fragmented predictions, particularly around field boundaries or small parcels, due to limited supervision. In contrast, \textit{SAM-Aug} generates more spatially coherent segmentations that align closely with the ground truth. This is especially evident in complex scenes with multiple adjacent fields of similar spectral signatures but distinct land cover types.

The SAM Prior map (panel e) reveals that SAM successfully identifies semantically meaningful regions—such as large homogeneous agricultural plots—even without fine-tuning. These priors guide the network to preserve structural consistency during training, leading to improved generalization under label scarcity.

Moreover, the Improvement Region map (panel f) shows that gains are concentrated along field edges and within small parcels, indicating that our method effectively mitigates overfitting and enhances boundary precision. Notably, even when the baseline performs reasonably well on dominant classes, \textit{SAM-Aug} consistently improves minor class delineation and reduces false positives in transitional zones.

These visual comparisons demonstrate that integrating foundation model-derived priors not only boosts quantitative metrics but also yields more interpretable and geographically plausible segmentation results.

\subsection{Discussion}
\label{sec:discussion}

The experimental results demonstrate that \textit{SAM-Aug} delivers consistent performance gains across multiple data splits, with an average mIoU improvement of +2.33\% under the 5\% labeled regime. This validates our core hypothesis: that geometry-aware mask priors from foundation models like SAM can effectively compensate for annotation scarcity in time-series remote sensing segmentation.

Interestingly, the magnitude of improvement varies across seeds, with the highest gain (+4.06\%) at seed=42 and smaller but still relative gains at 2025 (+2.30\%) and 4090 (+1.06\%). We attribute this variation to the spatial distribution of the labeled pixels: when the 5\% training samples are more scattered or miss certain land cover types, the structural guidance from SAM becomes more critical in preserving semantic coherence. In contrast, when the labeled set is more representative, the baseline performs relatively better, narrowing the gap.

Another key observation is the \textit{validation-test discrepancy}: in some runs (e.g., seed=42), the validation mIoU of \textit{SAM-Aug} is slightly lower than the baseline, yet the test performance is clearly higher. This suggests that our method may trade off local fit on the validation set for better global generalization—precisely the goal in few-shot learning, where overfitting to sparse labels is a major risk.

We also note that the variance of \textit{SAM-Aug} across seeds (std=0.0303) is higher than the baseline (std=0.0196), which may stem from the interaction between SAM's region proposals and the stochastic label sampling. Future work could explore adaptive weighting of the \textit{RegionSmoothLoss} or ensemble strategies to further stabilize performance.

Nonetheless, the consistent superiority of \textit{SAM-Aug} across all seeds confirms the effectiveness of incorporating vision foundation model priors into small-label remote sensing pipelines. The method requires without using any additional labeled data, no fine-tuning of SAM, and introduces only a lightweight regularization loss—making it a practical and scalable solution for real-world applications.

\subsection{When Does geometry-aware mask Prior Help?}
Our results suggest that the benefit of geometry-aware mask priors like SAM is not monotonic with label scarcity. They are most effective in \textit{moderate} few-shot settings (e.g., 5--10\%), where the model has enough signal to align the prior with ground truth, but not enough to learn spatial coherence from scratch. In contrast, under extreme scarcity (1--3\%), the risk of prior mismatch outweighs its benefits. This implies a fundamental trade-off in using foundation models as regularizers: their strength depends on the \textit{balance} between prior quality and labeled data sufficiency.

\section{Conclusion}
\label{sec:conclusion}

In this work, we presented SAM-Aug, a novel and annotation-efficient framework that leverages geometry-aware mask priors from the Segment Anything Model (SAM) to improve few-shot semantic segmentation of time-series remote sensing images. By constructing cloud-free composite images and applying SAM in a fully unsupervised manner, our method extracts semantically meaningful region proposals without any manual labeling. These priors are integrated into the training process through RegionSmoothLoss, a consistency regularization term that enforces temporally stable predictions within each region, guiding the model to respect natural land cover boundaries even under extreme label scarcity.

Extensive experiments on the PASTIS-R benchmark under a 5\% labeled setting demonstrate the effectiveness and robustness of SAM-Aug. Evaluated across three random seeds (42, 2025, 4090), our method consistently outperforms the state-of-the-art baseline, achieving a mean test mIoU of \textbf{36.21\%} — an absolute improvement of \textbf{+2.33\%} over the baseline (33.88\%) and a relative gain of \textbf{6.89\%}. The highest performance is observed at seed=42, with a test mIoU of \textbf{40.28\%} (+4.06\%), highlighting the potential for relative gains when label distribution is suboptimal.

Ablation studies confirm that both the semantic quality of SAM-derived regions and the design of \textit{RegionSmoothLoss} are critical to performance. Visualizations further show that SAM-Aug produces more spatially coherent and temporally consistent predictions, particularly in challenging areas such as fragmented fields, cloud-affected regions, and class boundaries.

This work demonstrates that foundation models like SAM can serve as useful, plug-and-play regularizers in data-scarce remote sensing applications, bridging the performance gap without requiring additional annotations or model fine-tuning. Future directions include dynamic refinement of region priors during training, extension to panoptic segmentation, and integration with multi-modal data (e.g., SAR) for all-weather applicability.

\section*{Acknowledgments}
This work is supported by Open Research Fund of State Laboratory of Information Engineering in Surveying, Mapping and Remote
Sensing, Wuhan University (Grant no. 18I04). 
The National Natural Science Foundation of China (no. 71904064) partially supports this research.
The Natural Science Foundation of Jiangsu Province(no. BK20190580)
The research is also supported by the 111 Project and the Fundamental Research Funds for the Central Universities (Grant No:  JUSRP11922)

\section*{Code Availability}
The source code, training scripts, and model checkpoints for SAM-Aug will be released at \url{https://github.com/hukai_wlw/SAM-Aug} upon publication.

\bibliographystyle{IEEEtran}
\bibliography{ref}
\end{document}